\documentclass{article}
\usepackage{ijcai26}
\usepackage{times}
\usepackage{soul}
\usepackage{url}
\usepackage[hidelinks]{hyperref}
\usepackage[utf8]{inputenc}
\usepackage[small]{caption}
\usepackage{graphicx}
\usepackage{amsmath}
\usepackage{amssymb}
\usepackage{booktabs}
\usepackage{algorithm}
\usepackage{algorithmic}
\usepackage[switch]{lineno}

\newtheorem{example}{Example}

\title{From Errors to Proofs: Minimal-Core-Guided Repair\\
for Neuro-Symbolic Constraint Solving}

\author{
    Dipankar Sarkar
    \affiliations
    Independent Researcher
    \emails
    dipankar@skelfresearch.com
}

\begin{document}

\maketitle

\begin{abstract}
Making language models solve constraint problems reliably often means having them translate the
problem into a formal specification and delegating the search to a sound solver. But the translation
is itself a language-model task, and an unfaithful translation makes the solver faithfully solve the
wrong problem. Existing pipelines repair only translations that crash, returning the solver's error
message and falling silent when the program runs but is wrong. We replace the error message with a
proof: when the generated program is unsatisfiable, we extract a minimal unsatisfiable core over the
model's own constraints and hand it back the exact set that cannot hold together, a leakage-free
signal that localizes the fault. On a new benchmark of 77 problems with an exact oracle, translation
to Answer Set Programming is faithful on six of seven domains and fails only on aggregate-coverage
scheduling, which concentrates the translation tax in one diagnosable pattern. A minimal core,
rather than a bare error, is what stops a weaker model from fabricating solutions to infeasible
problems, cutting fabrication from 79\% to 7\%. A strong chain-of-thought baseline meanwhile matches
the symbolic route on accuracy, so the route's value is not accuracy but certificates and its
refusal to fabricate.
\end{abstract}

\section{Introduction}
\label{sec:intro}

Language models are increasingly asked to solve discrete reasoning problems such as scheduling,
seating, assignment, and packing. Left to free-form generation they return fluent answers that
often violate the stated constraints, because autoregressive decoding has no mechanism to enforce
a global combinatorial constraint. A standard remedy separates translation from search: the model
converts the natural language problem into a formal specification, and a sound solver does the
reasoning \cite{pan2023logiclm,ye2023satlm,olausson2023linc,gao2023pal}. The solver supplies the
guarantee the model cannot.

This recipe is only as good as the translation, and language-model translations fail in two
different ways. Some are malformed and the solver reports a parse or grounding error. Others are
well formed but unfaithful: they add a constraint the problem never stated, which can make a
solvable problem report as unsatisfiable; they drop a stated constraint, which admits an invalid
solution; or they misstate the objective. The self-correction in current language-model plus
solver pipelines, exemplified by the self-refinement in Logic-LM \cite{pan2023logiclm}, reacts to
the first kind: it returns the solver's error message and asks the model to try again. When the
solver does not error, the loop is silent, and the well formed but unfaithful encoding survives.

We propose to repair with a proof rather than an error. We encode every decision as a single
\texttt{assign(Var,Value)} relation in Answer Set Programming
\cite{lifschitz2008asp,gebser2012aspbook} and solve with clingo \cite{gebser2019clingo}. When the
program is unsatisfiable, we do not report ``no solution''; we compute a minimal unsatisfiable
core over the model's own integrity constraints, the smallest subset that cannot all hold at once,
and hand that set to the model. The core is a proof artifact: it localizes the conflict to a few
constraints, which is exactly the information a generic error message withholds. It is computed
from the model's own program rather than from ground truth, so it adds no oracle leakage.

We then ask an empirical question the literature rarely tests: at the scales
where a solver can also certify the answer, how much does symbolic offloading actually help,
relative to plain prompting and to chain-of-thought, and where does it help. We answer it on a new
benchmark of 77 problems across seven domains with a programmatic oracle, on two open-weight
models. The answer is nuanced rather than a uniform win. Translation is
faithful on most domains and the symbolic route matches or beats chain-of-thought there, but a
single domain breaks it; a strong chain-of-thought baseline is hard to beat on accuracy; and the
distinctive payoff of the symbolic route is that it certifies optimality and infeasibility and
never fabricates a solution, which prompting cannot promise.

\footnotetext{Accepted as a poster at the IJCAI-ECAI 2026 Workshop on
Logic and Symbolic Reasoning (LogiSymb).}

\paragraph{Contributions.}
\begin{itemize}
\item \textbf{Minimal-core-guided repair} (Section~\ref{sec:approach}): we repair a language model's
formalization with a minimal unsatisfiable core over its own constraints rather than a solver error
message. The signal is structural and leakage-free, it ports the conflict-explanation idea from
constraint programming and answer set programming
\cite{junker2004quickxplain,gebser2008debugging} into the repair loop, and we isolate its effect with
a controlled error-message baseline \cite{pan2023logiclm}. The benefit is conditional: large on a
weaker model, negligible on a stronger one that rarely errs.
\item \textbf{A diagnosis of autoformalization fidelity} (Section~\ref{sec:results}): with an exact
oracle we show translation to ASP is faithful on six of seven domains and collapses only on
aggregate-coverage scheduling, localizing the translation tax to a formalization pattern rather than
the model.
\item \textbf{A recalibration of when offloading helps} (Sections~\ref{sec:results} and
\ref{sec:discussion}): a strong chain-of-thought baseline matches the symbolic route on accuracy, so
its value is certificates and robustness; the route never fabricates a solution for an infeasible
problem, whereas direct prompting does so 21\% of the time. We release the benchmark and oracle.
\end{itemize}

\section{Background and Related Work}
\label{sec:related}

Our work sits at the meeting point of four lines: tool-augmented reasoning, autoformalization,
self-correction, and conflict explanation. We take the method from the last and apply it to the
problem raised by the first three.

\paragraph{Faithful reasoning by offloading to solvers.}
A growing body of work makes language-model reasoning faithful by delegating inference to an
external system. Logic-LM \cite{pan2023logiclm} translates a problem into one of several symbolic
languages and calls a matching solver; SatLM \cite{ye2023satlm} generates a declarative
specification for a satisfiability-modulo-theories solver; LINC \cite{olausson2023linc} maps text
to first-order logic and calls a theorem prover; program-aided prompting \cite{gao2023pal} offloads
arithmetic to a Python interpreter. These methods target deductive question answering, on datasets
such as ProofWriter, FOLIO, and AR-LSAT, or arithmetic word problems. We target combinatorial
constraint satisfaction and optimization over Answer Set Programming, where correctness means
feasibility or optimality of an assignment rather than the truth of an entailment, and where the
relevant failure is silent over-constraint or under-constraint rather than a wrong proof step.

\paragraph{Autoformalization is the bottleneck.}
Once search is delegated, accuracy is bounded by the fidelity of the translation. Autoformalization
studies this translation directly, from natural language to formal mathematics
\cite{wu2022autoformalization}. For optimization specifically, the NL4Opt competition
\cite{ramamonjison2023nl4opt} and OptiMUS \cite{ahmaditeshnizi2024optimus} map word problems to
mixed-integer linear programs. Consistent with that line, we find that the bottleneck is the
formalization rather than the search, and that it is uneven across problem types: faithful on most
constraint domains, brittle on the one that needs aggregate-coverage reasoning. We also confirm a
known boundary, that Answer Set Programming is a poor vehicle for large numeric optimization, and
keep optimization instances small enough for the solver to certify optima.

\paragraph{Self-correction and the signal it uses.}
Iterative self-correction improves model outputs, and the methods differ mainly in what signal
drives the next attempt. Self-Refine \cite{madaan2023selfrefine} and Reflexion
\cite{shinn2023reflexion} use the model's own verbal critique; Self-Debugging
\cite{chen2024selfdebug} uses program execution results; Logic-LM \cite{pan2023logiclm} uses the
solver's error message. All of these react to an observed event: a critique, a failed test, a
crash. None uses a proof of why the current attempt is inconsistent. Our repair signal is exactly
such a proof, and our controlled comparison shows that what matters is not that repair happens but
what information it carries.

\paragraph{Conflict explanation and minimal unsatisfiable cores.}
Identifying a small set of constraints responsible for unsatisfiability is a classical problem in
constraint programming and satisfiability. Junker's QuickXplain \cite{junker2004quickxplain}
computes preferred minimal conflicts for over-constrained problems by divide and conquer, and
deletion-based filtering is the simplest member of that family. In Answer Set Programming the same
idea underlies debugging of inconsistent programs \cite{gebser2008debugging} and core-guided
optimization inside modern solvers \cite{gebser2019clingo}. Our contribution is to carry this
proof artifact across the boundary into the language-model repair loop: instead of asking the model
to debug from a flat error string, we hand it the minimal conflicting subset that a solver-side
conflict analysis produces. To our knowledge, minimal unsatisfiable cores have not been used as the
feedback signal for repairing a language model's formalization.

\section{Approach}
\label{sec:approach}

Our system follows the standard division of labor between a language model and a solver, but it makes
the translation step robust instead of assuming it correct. The model is responsible only for turning
a natural language problem into a formal specification; a sound solver does all of the search. Our
contribution is the loop that sits between them. When the solver reports that the
specification has no solution, we neither accept that verdict at face value nor restart the model
blindly. We extract a proof of the inconsistency, in the form of a minimal unsatisfiable core, and ask
the model to reconcile that proof with the problem text.

A problem passes through four stages. First, the model produces a structured ASP encoding over a
fixed decision relation. Second, the encoding is assembled into a program and run through clingo.
Third, the solver outcome is classified into one of four categories, and if it is not a clean success
a typed diagnostic is constructed from the solver state. Fourth, the diagnostic is returned to the
model, which revises the encoding; the second through fourth stages then repeat up to $K$ times. Only
the first stage uses the model creatively. The rest are deterministic and grounded in the solver, so
every correction the loop applies is justified by a concrete solver artifact rather than by another
free-form guess.

\subsection{The decision contract}
\label{sec:contract}
The model emits a JSON object with three fields: \texttt{facts}, which encode the problem data;
\texttt{rules}, which contain a choice rule that generates candidate decisions together with the
integrity constraints that enforce the requirements; and an optional \texttt{optimize} directive. We
impose a single contract on this output. Every decision is expressed through one relation,
\texttt{assign(Var,Value)}, in which each decision variable \texttt{Var} is bound to exactly one
\texttt{Value} by a choice rule, and every requirement is expressed as an integrity constraint, a rule
that begins with \texttt{:-} and forbids the combinations that violate it. The harness appends
\texttt{\#show assign/2.} and calls clingo. Figure~\ref{fig:encoding} shows a complete encoding for a
small graph-coloring instance.

\begin{figure}[t]
\centering
\begin{minipage}{0.98\columnwidth}
\scriptsize
\begin{verbatim}
% facts (problem data)
node(n1). node(n2). node(n3).
color(red). color(green). color(blue).
% choice rule: one colour per node
1 { assign(N,C) : color(C) } 1
     :- node(N).
% integrity constraints:
% adjacent nodes get different colours
:- assign(n1,C), assign(n2,C).
:- assign(n2,C), assign(n3,C).
% harness appends:  #show assign/2.
\end{verbatim}
\end{minipage}
\caption{A complete \texttt{assign/2} encoding of a three-node graph-coloring problem. The choice
rule generates one colour per node; each integrity constraint forbids an edge from being monochromatic.}
\label{fig:encoding}
\end{figure}

This contract serves three purposes, and each matters for the rest of the method. First, it
gives a single, uniform notion of a solution across every domain, so that one oracle can score the
output of any system, whether it is the symbolic pipeline or a prompting baseline, without
domain-specific parsing; a coloring, a roster, a seating, and a knapsack packing are all just sets of
\texttt{assign} atoms. Second, it keeps the encoding compact and stable in size as the problem grows,
because the decision logic lives in a constant number of rules while only the facts scale; a graph
with three nodes and a graph with ninety nodes share the same choice rule and the same constraint
schemas. Third, and most important for repair, it cleanly separates the part of the program the model
might get wrong, the integrity constraints, from the part that is fixed by the data, the facts and the
choice rule, which is exactly the separation the core extraction in Section~\ref{sec:repair} exploits.
To keep the comparison fair, every system in our study, including the prompting baselines, is given
the same decision variables and their allowed values, so no system is advantaged or penalized by the
output format and all of them are judged purely on whether they reason to a correct assignment.

\subsection{Compiling and classifying the outcome}
\label{sec:taxonomy}
Assembling and solving an encoding yields exactly one of four outcomes, and the loop reacts to each
differently. An outcome is \textsc{ok} when clingo returns an answer set that contains
\texttt{assign/2} atoms; this is the only terminating outcome, and the atoms become the assignment.
It is \textsc{syntax\_error} when grounding or parsing fails, which happens when the model writes a
malformed rule or an unsafe variable. It is \textsc{empty} when the program is satisfiable but
produces no \texttt{assign/2} atoms, which happens when the model forgets the choice rule or models
its decisions under some other predicate. It is \textsc{unsat} when the program is well formed and
grounds cleanly but has no answer set.

The four outcomes are not equally informative, and the asymmetry is the reason the rest of the method
exists. A \textsc{syntax\_error} is loud: clingo points at the offending rule, and prior error-message
refinement already handles this case well. The dangerous outcome is \textsc{unsat}, because it is
silent about its cause. An over-constrained translation of a perfectly solvable problem, for instance
one that hallucinates a constraint the text never stated, produces a program with no answer set, and
that program is indistinguishable, from the bare verdict alone, from a faithful translation of a
genuinely infeasible problem. The model that reads only ``no answer set'' cannot tell whether it
should fix its own constraints or report that the problem has no solution. Resolving that ambiguity is
what the core provides.

\subsection{Repair with a minimal unsatisfiable core}
\label{sec:repair}
On \textsc{unsat} we localize the conflict before asking the model to act. Let
$C=\{c_1,\dots,c_m\}$ be the integrity constraints the model produced and $B$ the base program made of
the facts and the choice rule. Because $B$ is satisfiable on its own (the choice rule alone always has
models), the inconsistency must come from a subset of $C$. We compute a minimal unsatisfiable subset
$C^\star\subseteq C$, defined by the property that $B\cup C^\star$ is unsatisfiable while
$B\cup(C^\star\setminus\{c\})$ is satisfiable for every $c\in C^\star$, so that every constraint in
$C^\star$ is necessary for the conflict and none is redundant. We obtain it by deletion filtering
(Algorithm~\ref{alg:core}), the simplest member of the QuickXplain family of conflict-extraction
procedures \cite{junker2004quickxplain}: starting from the full set, we tentatively remove each
constraint and keep the removal whenever the remainder is still unsatisfiable. The procedure issues at
most $m$ satisfiability checks, each a fast call on a small program, and for the instance sizes in our
benchmark its cost is negligible next to a single model call.

\begin{algorithm}[t]
\caption{Minimal unsatisfiable core (deletion filtering)}
\label{alg:core}
\begin{algorithmic}[1]
\REQUIRE base $B$, constraints $C$; $B\cup C$ unsatisfiable
\STATE $C^\star \gets C$
\FORALL{$c \in C$}
  \IF{$B \cup (C^\star \setminus \{c\})$ is \textsc{unsat}}
     \STATE $C^\star \gets C^\star \setminus \{c\}$ \COMMENT{drop $c$}
  \ENDIF
\ENDFOR
\RETURN $C^\star$
\end{algorithmic}
\end{algorithm}

Two properties of this signal matter. It is \emph{structural}: $C^\star$ is a proof
that those constraints cannot coexist, not a report that something went wrong, and it names the exact
constraints to reconsider rather than leaving the model to search its whole program. It is
\emph{leakage-free}: every satisfiability check in Algorithm~\ref{alg:core} runs on the model's own
program, with the objective removed, and at no point does the procedure consult the ground-truth
answer, the oracle, or the reference encoder. The core therefore tells the model where its translation
is internally inconsistent, which is information the model is entitled to, without telling it what the
correct answer is, which it is not. This distinction is what lets us use the core during inference
without contaminating the evaluation.

The repair prompt for \textsc{unsat} lists the constraints in $C^\star$ verbatim and frames the choice
the core makes explicit: either one of these constraints misformalizes the problem, in which case the
model should correct or remove it, or the named constraints all faithfully encode the text and the
problem is genuinely infeasible, in which case the model should leave the encoding unchanged so that
the system reports infeasibility. This is the central design point. By naming the conflict, the core
makes ``the problem is infeasible'' a legitimate and well supported answer, which is the
answer a model is likely to abandon when it is told only that the solver found nothing. The other two
failure types take simpler prompts: \textsc{syntax\_error} carries the grounder message and asks for a
parse fix, and \textsc{empty} asks the model to add the choice rule so that decisions surface as
\texttt{assign/2}. After any repair the program is re-solved, and the loop runs up to $K=3$ times.
Example~\ref{ex:core} traces the loop on a real instance from our benchmark.

\begin{example}[Repairing an over-constrained roster]
\label{ex:core}
On a six-worker, five-day roster, the model's first encoding is unsatisfiable. Deletion filtering
returns a minimal core of seven constraints: the coverage requirement for the final day together with
the six per-worker caps that, taken together, leave no worker free to staff that day. The repair
prompt lists exactly these seven constraints. The model recognizes that it had encoded the final-day
coverage more strictly than the text required, relaxes that one constraint, and the next solve returns
a valid roster, in a single repair iteration. A generic error message for the same instance would have
said only that the program had no answer set, and, as Section~\ref{sec:results} shows, a weaker model
given that message tends to keep editing until it returns a roster for a problem that has none.
\end{example}

\subsection{Isolating the signal: a faithful error-message baseline}
\label{sec:baseline}
To attribute any effect to the core rather than to the act of repairing, we build a second repair
setting that is identical in every respect except the feedback. It uses the same pipeline, the same
model, the same iteration budget, and the same decision contract, but on any failure it returns only
the raw solver message, with no core and no per-failure guidance. This reconstructs the error-message
self-refinement of prior language-model plus solver systems \cite{pan2023logiclm} inside our setup. We
refer to the two settings as \textsc{typed+core} and \textsc{generic}. Because they differ only in the
information content of the feedback, any difference in their behavior is attributable to the core, and
Section~\ref{sec:results} reports that the difference is small on a strong model and large on a weak
one.

\section{Benchmark}
\label{sec:benchmark}

We generate 77 instances over seven domains: graph coloring, knapsack, team assignment, seating,
timetabling, duty scheduling, and Latin squares. Each instance pairs a description with a structured
specification over \texttt{assign/2}; the description states every constraint in the specification,
so a system is never penalized for information it was not given. A deterministic encoder,
independent of the language model, computes ground-truth feasibility and, for optimization
instances, the optimum; a separate oracle scores any assignment for constraint satisfaction,
feasibility, and objective. Instances carry one of five labels: \emph{feasible}, \emph{tight}
(numerically tight but solvable), \emph{infeasible} (the correct behavior is to report
unsatisfiability), \emph{stress} (larger), and \emph{xl} (60 to 120 decision variables). The xl tier
is feasibility-only, because clingo certifies one model quickly at that size but cannot certify
large numeric optima or large infeasibility quickly.

\section{Experimental Setup}
\label{sec:setup}

\paragraph{Models.} We use two open-weight models, \texttt{gemma4:31b} as the primary model and
\texttt{deepseek-v3.2}, served through a hosted inference provider with greedy decoding. Both are
run with the same pipeline and prompts.

\paragraph{Systems.} (B1) direct: the model outputs the assignment. (B2) chain-of-thought
\cite{wei2022cot}: reason then answer. (B3) the ASP pipeline with no repair. (B4) the pipeline with
Logic-LM-style generic-error repair. (B5) the pipeline with typed plus minimal-core repair.

\paragraph{Metrics.} Feasibility accuracy (a valid assignment on feasible instances, or a correct
unsatisfiability report on infeasible ones); constraint-satisfaction rate; fabricated feasibility,
a committed solution for an infeasible problem; optimality gap on optimization instances; and
average repair iterations. We omit per-system token counts, because shared response caching across
the three pipeline settings confounds them.

\section{Results}
\label{sec:results}

Table~\ref{tab:main} reports the primary model over all 77 instances.

\begin{table}[t]
\centering
\small
\setlength{\tabcolsep}{4.5pt}
\begin{tabular}{lrrrr}
\toprule
System & Feas.\ acc. & CSR & Fabr. & Iters \\
\midrule
Direct                 & 74.0 & 84.5 & 21.4 & 0.00 \\
Chain-of-thought       & \textbf{98.7} & \textbf{99.8} & 0.0 & 0.00 \\
ASP, no repair         & 84.4 & 85.0 & 0.0 & 0.00 \\
ASP, generic repair    & 88.3 & 91.3 & 0.0 & 0.78 \\
ASP, typed+core repair & 89.6 & 91.3 & 0.0 & 0.87 \\
\bottomrule
\end{tabular}
\caption{Primary model (\texttt{gemma4:31b}), 77 instances, percent. Higher is better except
fabricated feasibility (Fabr.), measured over the 14 infeasible instances.}
\label{tab:main}
\end{table}

\paragraph{Direct prompting is unreliable; chain-of-thought is strong.}
Direct prompting reaches 74.0\% feasibility accuracy and fabricates a solution on 3 of 14 infeasible
instances (21.4\%). Chain-of-thought reaches 98.7\% and fabricates none. At the scales in this
benchmark, an explicit reasoning chain solves, optimizes, and recognizes infeasibility for almost
every instance without any solver.

\paragraph{The translation tax is concentrated in one domain.}
The pooled pipeline numbers understate what is happening per domain. Table~\ref{tab:domain} shows
that on six of seven domains the no-repair pipeline already reaches 100\% feasibility accuracy, and
on coloring it exceeds chain-of-thought (100\% versus 94.7\%). The entire deficit comes from
scheduling, where the model must encode aggregate coverage and per-worker caps with counting
constraints, and where the no-repair pipeline scores 0\%. In other words, language-model
translation to ASP is faithful wherever the constraints are local, and breaks on the one domain that
needs global counting. This is a sharper statement than ``the pipeline is less accurate'': it tells
us which formalization patterns the model can be trusted to produce.

\begin{table}[t]
\centering
\begin{tabular}{lrrr}
\toprule
Domain & CoT & ASP no rep. & ASP +core \\
\midrule
graph coloring   & 94.7 & \textbf{100.0} & \textbf{100.0} \\
knapsack         & 100.0 & 100.0 & 100.0 \\
Latin square     & 100.0 & 100.0 & 100.0 \\
seating          & 100.0 & 100.0 & 100.0 \\
team assignment  & 100.0 & 100.0 & 100.0 \\
timetabling      & 100.0 & 100.0 & 100.0 \\
scheduling       & 100.0 & 0.0 & 33.3 \\
\bottomrule
\end{tabular}
\caption{Feasibility accuracy by domain, primary model. The pipeline matches or beats
chain-of-thought everywhere except scheduling, which needs aggregate-coverage encoding.}
\label{tab:domain}
\end{table}

\paragraph{Repair is monotone and a core helps where it is needed.}
Self-repair lifts the pipeline from 84.4\% to 89.6\% on the primary model. The improvement is safe:
across all 77 instances, repair turned four first-attempt failures into successes and broke none.
Fifty-four instances solved on the first attempt; the rest entered the loop, and the instances that
exhausted the budget are exactly the scheduling cases the model cannot encode. On the primary model,
typed plus minimal-core repair (89.6\%) edges generic repair (88.3\%), but the gap is within noise
for a model this strong (paired sign test $p=1.0$): a model that rarely produces a spurious
unsatisfiability rarely needs the core. The difference is stark on a weaker model. On
\texttt{deepseek-v3.2}, generic error-message repair fabricates a solution on 11 of 14 infeasible
instances (79\%), because, told only that the program has no answer set, it deletes a real
constraint until the solver returns a model; minimal-core repair fabricates on 1 of 14 (7\%), a
difference that is significant by a two-sided Fisher exact test ($p<10^{-3}$,
Table~\ref{tab:multimodel}). Appendix~\ref{app:trace} traces both behaviors on one infeasible
instance. Naming the conflicting constraints is what prevents the model from fabricating a
solution.

\paragraph{Certificates and robustness.}
No pipeline setting ever fabricates a solution for an infeasible problem (0\% in Table~\ref{tab:main}),
against 21.4\% for direct prompting, because clingo either returns an answer set or proves there is
none. On optimization instances the returned solution is provably optimal (zero optimality gap),
whereas a prompted answer is optimal only by inspection. These are guarantees that a prompting
system, however accurate in aggregate, cannot make about any single answer.

\begin{figure}[t]
\centering
\includegraphics[width=\columnwidth]{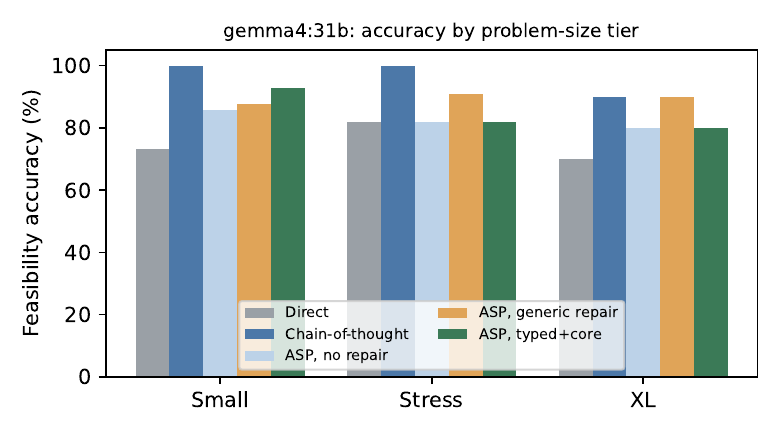}
\caption{Feasibility accuracy by problem-size tier on the primary model
(small $n{=}56$, stress $n{=}11$, xl $n{=}10$). The pipeline beats direct prompting throughout but
does not overtake chain-of-thought, whose only visible drop is on the xl tier.}
\label{fig:tier}
\end{figure}

\paragraph{Scale.}
Figure~\ref{fig:tier} breaks the primary model down by problem-size tier. Chain-of-thought stays at
or near 100\% through the stress tier and drops only to 90\% on the xl tier. The clearest accuracy
crossover is a single xl instance, a 90-node three-colorable graph, where chain-of-thought violates
an edge while the core pipeline returns a valid coloring. The pipeline has its own scale failure on
a 120-variable scheduling instance, the same domain that fails at small scale. At the sizes a solver
can certify quickly, neither method dominates the other on accuracy.

\begin{table}[t]
\centering
\begin{tabular}{lrrr}
\toprule
Model & Direct & Generic repair & Core repair \\
\midrule
\texttt{gemma4-31b} & 21 & 0 & 0 \\
\texttt{deepseek-v3.2} & 0 & 79 & 7 \\
\bottomrule
\end{tabular}

\caption{Fabricated feasibility, the rate at which a system commits to a solution for an
infeasible problem, over the 14 infeasible instances, by model and repair signal. Generic
error-message repair drives a weaker model to fabricate; a minimal core does not.}
\label{tab:multimodel}
\end{table}

\section{Discussion}
\label{sec:discussion}
We draw three lessons. First, offloading constraint reasoning to a solver should be justified by
guarantees and robustness, not by accuracy alone: a strong
chain-of-thought baseline is hard to beat on accuracy at the scales where a solver can also certify
the answer, but it cannot certify infeasibility or optimality, and direct prompting fabricates
feasibility often enough (21\% here) to matter wherever a wrong answer is costly. Second, the
fidelity of language-model autoformalization is uneven and the unevenness is structured: it is
near-perfect for local constraints and breaks on global counting, which suggests that the right unit
of trust is the formalization pattern, not the model. Third, the form of the repair signal matters
most when the model is weak; a minimal core tells the model which constraints conflict, which
discourages the fabrication that a flat error message can induce. The broader message for symbolic
reasoning with language models is that solver-side artifacts, not just solver verdicts, are useful
feedback, and minimal cores are one such artifact.

\section{Limitations}
\label{sec:limitations}
The benchmark uses templated descriptions and a single \texttt{assign/2} decision relation, which
fits discrete constraint and optimization problems but not problems with derived numeric quantities.
Answer Set Programming cannot certify large numeric optima or large infeasibility quickly, which
bounds how far we can push scale while keeping exact ground truth, and the accuracy crossover is
therefore a single instance. Repair is driven by solver-intrinsic signals only, so a missing
constraint that yields a satisfiable but invalid encoding is caught by the oracle at evaluation
rather than repaired in the loop; a verifier in the loop is the natural next step.

\paragraph{Threats to validity.}
Four caveats bound our conclusions. First, the benefit of the core over a bare error message is
demonstrated on one of our two models; on the stronger model the two are indistinguishable, so the
result identifies a regime, weaker and fabrication-prone models, in which feedback shape matters,
rather than a universal gain. Confirming the effect on more models is the most important next step,
which we leave to future work. Second, every system is given the decision variables and their
allowed values, which makes the task easier than open-ended translation and, if anything, flatters
the prompting baselines and understates the symbolic route's relative value. Third, the symbolic
pipeline issues several model calls per problem against one or two for prompting, so its accuracy
parity comes at a compute premium that we do not chart in detail. Fourth, the scheduling failure is
reported under a fixed prompt, and a different prompt or a worked counting example might narrow it,
so the per-domain diagnosis should be read as a property of this configuration rather than a law of
autoformalization.

\section{Conclusion}
\label{sec:conclusion}
We replaced the error message in a language-model plus solver repair loop with a proof: when the
generated program is unsatisfiable, we hand the model a minimal unsatisfiable core over its own
constraints. On 77 constraint and optimization problems and two open-weight models, the resulting
pipeline is faithful on six of seven domains, repairs monotonically, and uses the core exactly where
weaker models would otherwise hallucinate feasibility. The same study shows that symbolic offloading
does not beat a strong chain-of-thought baseline on accuracy at verifiable scales; its value is that
it never fabricates a solution and certifies the answers it does return. We read these results as an
argument to treat solver-side proof artifacts as first-class feedback for language-model
formalization, and we release the benchmark and oracle to make the scale question sharper.

\appendix
\section{Repair feedback and a worked trace}
\label{app:trace}

The two repair settings share the pipeline, the model, and the iteration budget, and differ only in
the feedback returned on an \textsc{unsat} outcome. The generic setting returns the raw verdict:

{\scriptsize
\begin{verbatim}
clingo found no answer set (unsatisfiable).
\end{verbatim}
}

\noindent The typed-core setting returns the minimal core and the decision it implies:

{\scriptsize
\begin{verbatim}
The following integrity constraints are JOINTLY
UNSATISFIABLE (a minimal conflicting set):
  - :- assign(X,C),assign(Y,C),border(X,Y).
If one of them misformalizes the problem,
correct or remove it; if the problem is
genuinely infeasible, leave it unchanged.
\end{verbatim}
}

\noindent\textbf{Trace on an infeasible instance.} The instance is a four-node clique, every pair of
nodes adjacent, with three colors, which is genuinely infeasible because a clique of four needs four
colors. On \texttt{deepseek-v3.2}:

\begin{itemize}
\item \emph{Generic repair.} The first solve is unsatisfiable and the model receives only ``no
answer set.'' It responds by weakening its coloring constraint, which makes the program
satisfiable, and returns \texttt{assign(n3,c1)}, \texttt{assign(n4,c1)}, \texttt{assign(n2,c2)},
\texttt{assign(n1,c3)}. Nodes \texttt{n3} and \texttt{n4} are adjacent yet share color \texttt{c1}:
the system reports an invalid coloring as a solution.
\item \emph{Minimal-core repair.} Each solve is unsatisfiable, and each time the feedback names the
single coloring constraint as the conflict. The model judges that constraint to be a faithful
encoding of adjacency, leaves it unchanged, and after the iteration budget reports the instance
infeasible, which is correct.
\end{itemize}

\noindent This is the mechanism behind the aggregate numbers: a bare verdict invites the model to
dissolve the conflict by dropping a true constraint, while the core makes ``the problem is
infeasible'' the better-supported response.

\bibliographystyle{named}
\bibliography{refs}

\end{document}